\documentclass[10pt,journal,compsoc]{IEEEtran}

\ifCLASSOPTIONcompsoc
  \usepackage[nocompress]{cite}
\else
  \usepackage{cite}
\fi
\ifCLASSINFOpdf
\else
\fi
\usepackage{amsmath,amssymb,amsfonts}
\usepackage{algorithmic}
\usepackage{graphicx}
\usepackage{textcomp}
\usepackage{xcolor}
\usepackage{verbatim}
\usepackage{booktabs}
\usepackage{caption}

\usepackage[pagebackref,breaklinks,colorlinks,bookmarks=false]{hyperref}
\usepackage{float}
\usepackage{titlesec}

\newcommand{\fig}[4]{
\begin{figure}[htp]
    \centering
    \includegraphics[width=#1mm,scale=1]{imgs/#2}
    \caption{#3}
    \label{fig:#4}
\end{figure}}
\begin{document}

\newcommand{\AFpaper}{$\text{(AF)}^2\text{-S3Net}$}

\def\sectionautorefname{Section}%
\let\subsectionautorefname\sectionautorefname
\let\subsubsectionautorefname\sectionautorefname

\author{\textit{Niklas Eisl, Dan Halperin}}

\title{\huge Point Cloud Based Scene Segmentation: A Survey}


\IEEEtitleabstractindextext{%
\begin{abstract}
Autonomous driving is a safety-critical application, and it is therefore a top priority that the accompanying assistance systems are able to provide precise information about the surrounding environment of the vehicle. Tasks such as 3D Object Detection deliver an insufficiently detailed understanding of the surrounding scene because they only predict a bounding box for foreground objects. In contrast, 3D Semantic Segmentation provides richer and denser information about the environment by assigning a label to each individual point, which is of paramount importance for autonomous driving tasks, such as navigation or lane changes. To inspire future research, in this review paper, we provide a comprehensive overview of the current state-of-the-art methods in the field of Point Cloud Semantic Segmentation for autonomous driving. We categorize the approaches into projection-based, 3D-based and hybrid methods. Moreover, we discuss the most important and commonly used datasets for this task and also emphasize the importance of synthetic data to support research when real-world data is limited. We further present the results of the different methods and compare them with respect to their segmentation accuracy and efficiency.
\end{abstract}

}


\maketitle

\IEEEdisplaynontitleabstractindextext

%
\IEEEpeerreviewmaketitle

\section{Introduction}

%
%
%
%

\IEEEPARstart{S}{emantic} Segmentation plays a critical role in providing autonomous vehicles with a sophisticated understanding of driving scenes to eventually deploy them in the real world. The goal is to assign a semantic class label to each basic component in a given environment, i.e., each pixel in a 2D image or each point in a 3D point cloud. Early research in this area was based on image data that relied solely on camera sensors. However, by projecting data from the 3D physical world onto the image plane, valuable information, such as depth, is lost \cite{zhou2020cylinder3d}. Therefore, the focus has shifted to the use of 3D LiDAR point cloud data \cite{LidarAD}. In addition, cameras cannot reliably capture data with rich information at night or in difficult weather conditions, while LiDAR sensors are much more robust in these challenging settings.


However, applying approaches that are based on LiDAR data to autonomous vehicles remains challenging due to the complexity of 3D data in terms of memory consumption and increased inference time. Furthermore, the density of the point cloud varies substantially depending on the distance to the LiDAR sensor, i.e. points positioned further away from the sensor are very sparsely distributed \cite{yan2020sparse}, which is a major issue especially for outdoor-driving scenarios. Unlike 2D Semantic Segmentation, where the algorithm processes dense inputs that are divided into regular pixels, an unordered set of points from a LiDAR scan is more difficult to handle. Moreover, there exist only a few large-scale datasets for 3D Semantic Segmentation, such as SemanticKITTI \cite{behley2019semantickitti} and nuScenes \cite{caesar2020nuscenes}.
Nevertheless, researchers have recently published several promising works to tackle these obstacles. There already exists a review paper \cite{survey} that presents a broad overview over numerous approaches in the area of Point Cloud Based Scene Segmentation, which spans the period from 2017, when the first work in this field started, until early 2020. However, to the best of our knowledge, there is no paper that covers the most recent approaches. 

In this review paper, we close this gap while focusing exclusively on the best performing methods, which are illustrated in \autoref{fig:Overview}. Thereby, we provide an in-depth overview of the current state-of-the-art approaches in 3D Point Cloud Segmentation.

\fig{87}{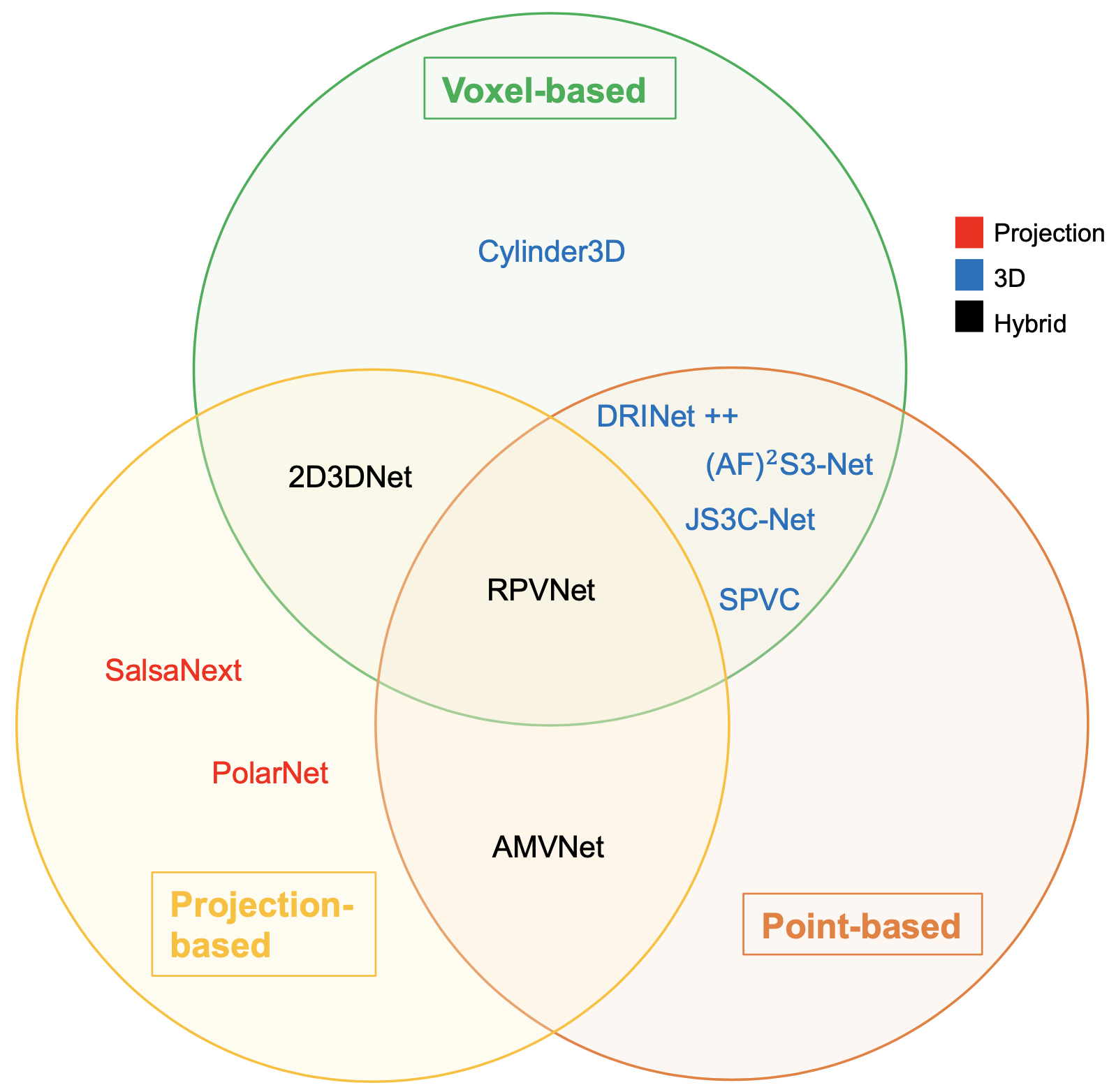}{Pioneering works can be categorized as either projection-based, voxel-based, or point-based. However, recent approaches typically combine different representations.}{Overview}

\vspace{-5mm} 
\section{Foundations}
In this section, we  introduce in detail important concepts that are used by the methods introduced in \autoref{sec:sota}.

\subsection{Sparse Convolutions and Submanifold Sparse Convolutions}

In 2D Image Processing, regular dense convolutional filters are commonly used due to the dense nature of pixels in images. However, in 3D, point clouds are sparsely distributed, so dense convolutions cannot extract meaningful features from the generated voxels. Moreover, applying dense convolutions in 3D space introduces an additional dimension that significantly increases the computational complexity and thus the inference time. Therefore, sparse convolutions \cite{choy20194d} only compute the convolutional operation for the active input sites, i.e. the nonzero cells. The data is stored more efficiently within a hash table that contains the values of the active sites and their positional information. However, this approach still suffers from the dilation problem, where deactivated neighboring cells of an active input site become non-zero due to the convolution operation. Therefore, submanifold sparse convolutions \cite{graham2017submanifold} extend the previous concept by specifying that a cell in the output feature map can only be activated if the corresponding input site is non-zero.

\subsection{Self-attention Mechanism} 
This concept \cite{attention} was first introduced in the field of Natural Language Processing (NLP), but is nowadays also extensively used in Computer Vision. The self-attention mechanism is a learned weighting over all elements of a sequence. Every embedding is transformed to a query vector $\mathbf{q}$ and a key vector $\mathbf{k}$, both of dimension $d_k$, and a value vector $\mathbf{v}$ of dimension $d_v$. The output is computed by
\begin{equation}
    Attention(\mathbf{q}, \mathbf{k}, \mathbf{v}) = softmax(\frac{\mathbf{q}\mathbf{k}^T}{\sqrt{d_k}})\mathbf{v}.
\end{equation}
The scores are calculated based on the dot-product between $\mathbf{k}$ and $\mathbf{q}$, and the softmax function converts them into probability-like weights, that determine how much an embedding should consider other parts of the sequence. Instead of considering all values $\mathbf{v}$ of the sequence equally, the values corresponding to higher scores contribute more to the output. 

\subsection{Generative Adversarial Networks (GANs)}
The architecture of a GAN \cite{goodfellow2014generative} consists of a generator $G$ and a discriminator $D$, which are two neural network models that compete against each other. The generator $G$ takes a latent space of noise $\mathbf{z}$ as input and learns the distribution over a dataset in such a way that it is able to reproduce samples $G(\mathbf{z})$ that could belong to the original dataset. The discriminator $D$ receives samples from both the generator $G(\mathbf{z})$ and samples $\mathbf{x}$ from the original training dataset. Then, it distinguishes between the two by classifying them as either real ($D(\mathbf{x})) = 1$) or fake ($D(G(\mathbf{z})) = 0$). The generator tries to mislead the discriminator and bring $D(G(\mathbf{z}))$ close to $1$ while the latter tries to keep $D(G(\mathbf{z}))$ close to $0$, when it is presented with a fake sample from the generator. Competition helps both these networks to enhance their performance.

\subsection{Feature Representations for Learning on Point Clouds}
Early methods have usually been categorized into projection-based, voxel-based or point-based methods, as illustrated in \autoref{fig:Overview}.

\subsubsection{Projection-based methods} \label{sec:proj_based}

These methods project the 3D LiDAR point cloud onto a 2D plane and then project the processed data back onto the 3D domain to obtain the segmented 3D scene. In doing so, they are highly efficient in terms of runtime because they require fewer parameters and can take advantage of low-cost 2D convolutions for fast GPU computation. However, due to the loss of information, these methods do not perform as well as methods that apply 3D operations. 

\subsubsection{Voxel-based methods} \label{Voxel-based methods}
Voxel-based methods aim to apply structure to unstructured point cloud inputs. 
To this end, the point cloud is mapped into a grid similar to the 2D Cartesian coordinate grid used in image processing.
 Instead of pixels, however, the grid is composed of 3D cubes, typically referred to as voxels, which usually have an identical volume.
In the naive approach, the value of a voxel is calculated by taking the average over the intensities of the points that lie within it. By choosing the resolution of the grid, i.e. the number of voxels into which the volume is divided, one can determine how many points are considered for the intensity calculation of each voxel. 

Voxel-based methods typically outperform projection-based methods because they do not discard 3D information. However, they also suffer from a certain kind of information loss. The higher the resolution, the lower the spatial and geometrical losses are, although the load on the memory is heavier. Due to hardware limitations, that do not allow the use of arbitrarily large convolutional kernels, networks typically perform aggressive down-sampling operations to maintain acceptably large receptive fields, which are essential for global context information. Sparse convolutions mitigate this issue by maintaining a higher resolution, but still experience information loss. As a consequence, smaller objects are not properly captured by the coarse resolution, as illustrated in \autoref{fig:ResolutionFig}. Moreover, if points from different semantic classes fall within the same grid cell, they will be assigned to the same class label.

\fig{85}{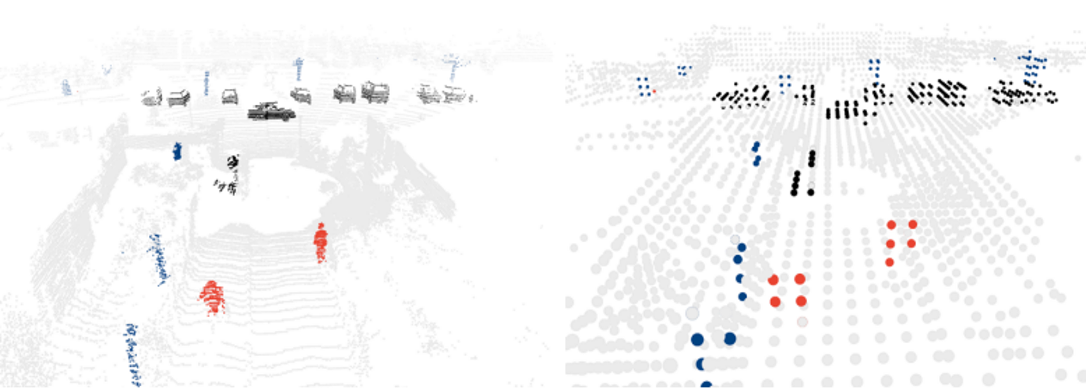}{\cite{tang2020searching} Smaller objects are no longer distinguishable at a low resolution. The left image shows a fine-grained 3D scene, while the right image has a coarse voxel resolution of $0.8\times 0.8\times 0.1$ [meters].}{ResolutionFig}

\subsubsection{Point-based methods}
Unlike the previous methods, point-based models do not introduce quantization errors, by directly operating on the raw point cloud. This way, geometric properties such as depth and height are preserved for each point.
Since  the point  cloud  is unordered and irregularly distributed, standard convolutions cannot be applied to it. To achieve permutation invariance, PointNet \cite{pointnet} processes every point independently with an MLP and extracts global features by a max-pooling operation. One problem of processing points individually is that point features cannot capture local context information from neighboring points. Works like PointNet++ \cite{pointnet++} address this issue by dividing the set of points into local regions. However, in contrast to voxelization, these regions overlap. To increase the receptive field, the extracted local features are aggregated into larger sets and further processed. To this end, most works perform a laborious search for local neighbors, which increases runtime due to the unstructured nature of the point cloud and makes these methods impractical for real-time applications. 
Point-based architectures such as PointNet can also be applied for voxelization. Instead of naively averaging over all points that fall within a voxel, one can also use learnable weights to extract meaningful voxel features.

\subsection{Dilated convolutions} \label{dilated}
In Semantic Segmentation, down-sampling operations are frequently used to increase the receptive field with an acceptable number of parameters. However, it is important to maintain a high resolution to obtain pixel-level or point-level accuracy, as described in \autoref{Voxel-based methods}.
Dilated convolutions \cite{dilated} are an effective alternative to obtain a large receptive field while maintaining high resolution at the same time. The dilated convolution $\ast_l$ between a discrete signal $F$ and a discrete kernel $k$ is defined as
\begin{equation}
    (F \ast_l k)(t):=\sum_{-\infty}^{\infty} F(t-l\tau) k(\tau),   
\end{equation}
where $l$ represents the dilation factor. In contrast to the standard convolution, the kernel $k$ touches $F$ only at each $l$-th entry. Therefore, increasing $l$ allows for larger receptive fields, while the number of parameters in the kernel remains the same. As the network depth grows, the number of parameters increases linearly, while the size of the receptive field rises exponentially, allowing for wider coverage.

%
%
%
%
\section{Point Cloud Semantic Segmentation} \label{sec:sota}

In Point Cloud Based Scene Segmentation, we consider a LiDAR scan with $N$ points, which can be represented as $\{(\mathbf{x}_i,\mathbf{y}_i) \, \vert \, i \in {1,...,N} \}$. The information from the unordered set of points consists of the 3D coordinates and the intensity of the laser beam $l \in \mathbb{R}$. More specifically, each input point can be expressed as $\mathbf{x}_i = \{(x,y,z),l \}$ coupled with the corresponding ground truth $\mathbf{y}_i$ that is represented as a one-hot encoded class label.
The goal of the task is to learn a parameterized function $\hat{\mathbf{y}}_i = f_\Phi(\mathbf{x}_i)$ that assigns a normalized class prediction to each point, such that the difference between $\hat{\mathbf{y}}_i$ and the ground truth $\mathbf{y}_i$ is minimized.

In the following, we introduce the current state-of-the-art approaches in the field of Point Cloud Semantic Segmentation. In this context, not only the segmentation performance is relevant, but also the inference time, which is of great significance for the task of autonomous driving.
We categorize the subsequent methods as projection-based, 3D-based or hybrid, as depicted in \autoref{fig:Overview}.

\subsection{Projection-based methods}

\subsubsection{Bird's-eye-view projection}

This type of projection represents the LiDAR scan by taking a top-down snapshot of the point cloud. In doing so, one looses the height information by neglecting the z-coordinate. 

\vspace{3mm} 
\textbf{PolarNet} \cite{zhang2020polarnet} is the current state-of-the-art method among all approaches that rely solely on bird's-eye-view projection (BEV). The authors have observed that the regular grid partitioning distributes the points in a non-uniform manner, and therefore abandons the inherent distribution of real-world 3D point clouds. Grid cells located close to the sensor contain significantly more points in each cell compared to cells that are further away from the sensor. To account for the ring-like structure of the point distribution, the authors propose to partition the LiDAR scan into a polar grid, which leads to a more efficient use of computational power and improved feature representativeness. Instead of using a Cartesian coordinate system, the input points are quantized based on their azimuth and radius, as illustrated in \autoref{fig:PolarGrid}. In order to obtain fixed-length representations for every grid, the authors suggest transforming the raw point data with PointNet \cite{pointnet}. 

\fig{80}{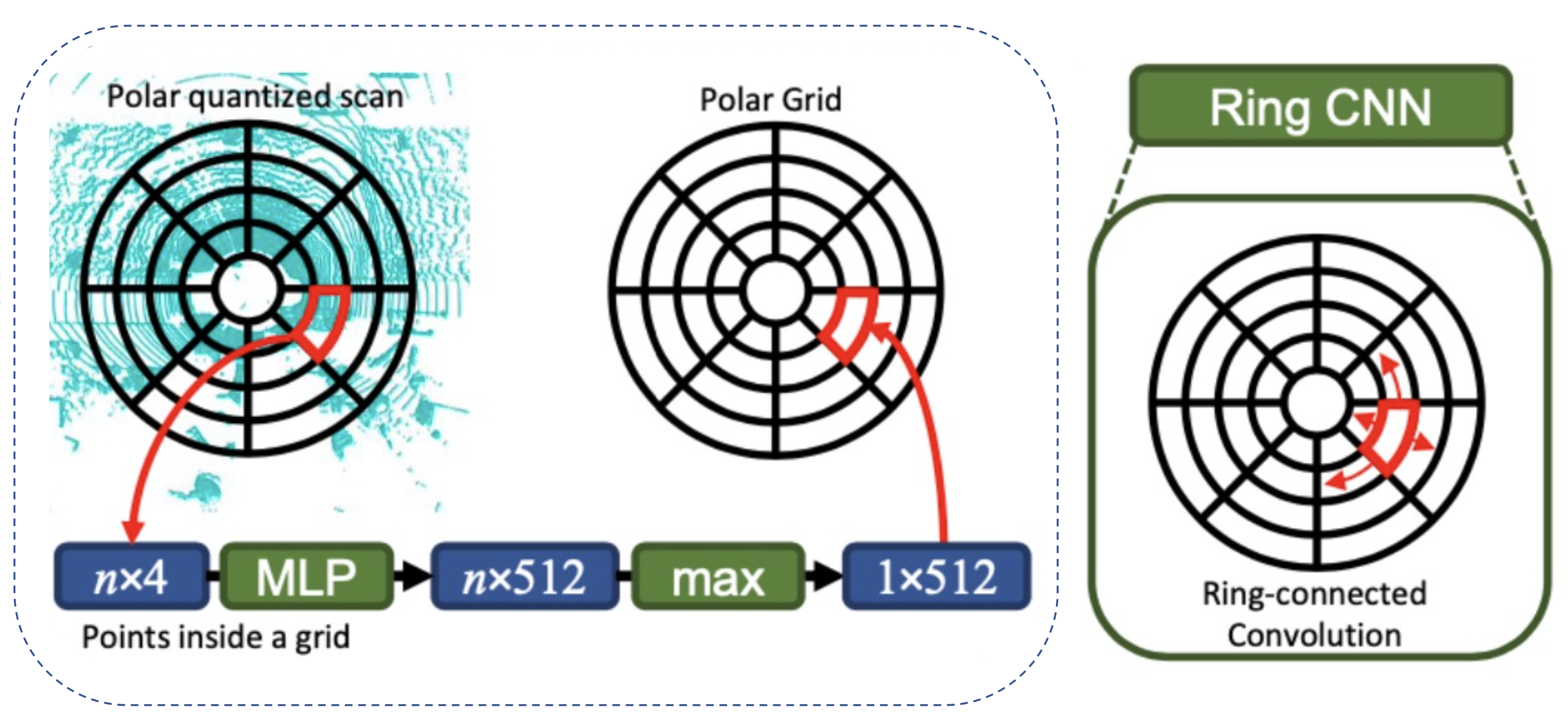}{\cite{zhang2020polarnet} A single grid cell consists of $n$ 4-dimensional points (3D coordinates and intensity), that are all independently processed with a PointNet to $n$ 512-dimensional representations. This is followed by a max-pooling operation to ensure that all grid cells have the same feature size of $1\times 512$.}{PolarGrid}

Furthermore, the authors introduce the Ring CNN, where the kernel shape adapts to the polar structure of the grid as depicted in \autoref{fig:PolarGrid}. Still, a ring convolutional operation works the same way as the standard CNN, where the kernel is rectangular according to the shape of an image. 
However, different from images where the left and right edges are not logically related to each other, information in circular point cloud data continuously depends on each other. It is therefore important to connect the two sides of the feature matrix along the azimuth-axis to allow for a convolution around the zero degree boundary.

\subsubsection{Range-view projection}
\fig{70}{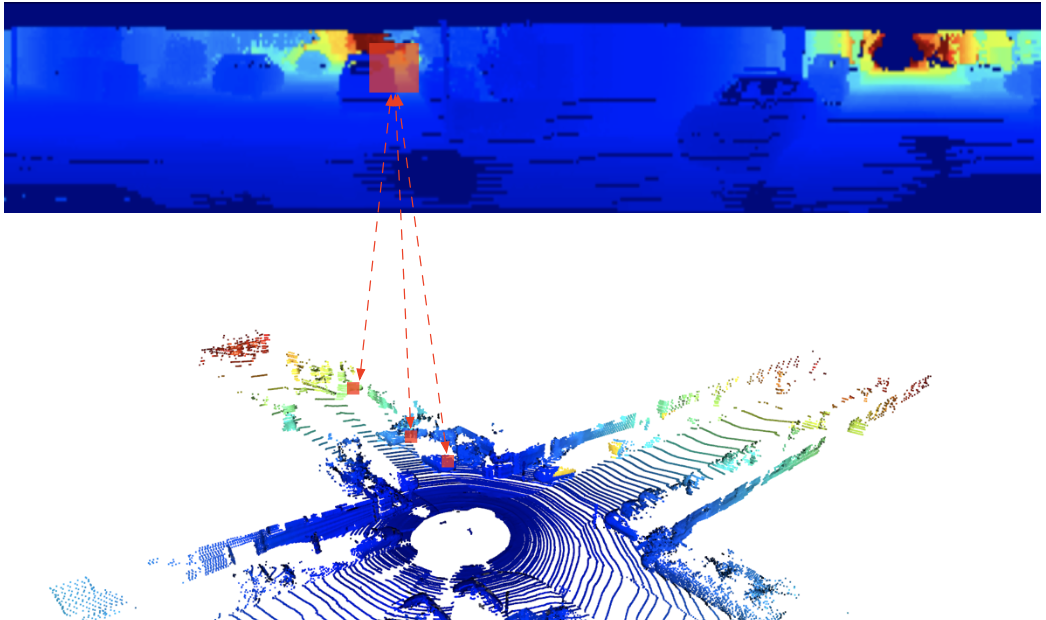}{\cite{zhou2020cylinder3d} The range-view projection maps the 3D points onto a 2D plane (panoramic view). Although the three red points have significantly different depth coordinates, they are all projected onto the same pixel,  resulting in a substantial loss of information.}{RvProj}

Another method to represent the 3D point cloud on a plane is to apply range-view projection (RV). The 3D LiDAR scan is mapped to a spherical panoramic view that encapsulates the entire $360^\circ$ field-of-view, which is illustrated in \autoref{fig:RvProj}. The spherical projection is performed using the \textit{atan2} and \textit{arcsin} functions. The process results in an image of shape $h\times w\times 5$, where $h$ and $w$ are the height and the width of the 2D plane. Each pixel stores the original ($x, y, z$) coordinate values, along with $l$ for the intensity and $r$, which represents the range of the corresponding point.

\vspace{3mm} 
\textbf{SalsaNext} \cite{cortinhal2020salsanext} is considered as the state-of-the-art approach among all networks that solely rely on RV-projection, and the architecture is heavily inspired by SalsaNet \cite{salsa}. Several key design decisions help to significantly increase segmentation performance while maintaining real-time throughput.
The network receives a 5-dimensional RV-image as input, follows an encoder-decoder architecture with residual skip-connections and finally outputs a 3D segmented point cloud.

To keep the model efficient, the authors reduce the number of parameters. For this purpose, a novel Pixel-Shuffle Layer is used for up-sampling instead of transposed convolutions in the decoder. This layer rearranges the feature map from $(H\times W\times Cr^2)$ to $(Hr\times Wr\times C)$, where $r$ is the upscaling ratio, $H$ and $W$  are the height and the width of the old feature map and $C$ is the channel size of the new feature map. A major advantage over other up-sampling methods that do not use learnable weights is that it does not produce checkerboard artifacts. For down-sampling in the encoder, a simple average pooling operation is applied instead of strided convolutions.

To improve the segmentation performance, the authors adopt several design choices. Firstly, they exploit the advantages of dilated convolutions, which are discussed in more detail in  \autoref{dilated}. At the beginning of the architecture, a novel Contextual Module is introduced to capture global context information by fusing receptive fields at multiple scales. To this end, the authors use $1\times1$ and $3\times3$ kernels with different dilation rates of $1$ and $2$. Additionally, each encoder block performs feature extraction with a series of layers that apply dilated convolutions with effective receptive fields of 3, 5, and 7. The extracted features are concatenated and further processed with a $1\times1$ convolution.
Moreover, an approach called Central Encoder-Decoder Dropout is used to ensure that dropout is not applied to the first and last layers of the architecture, but only to the central layers. The reason behind this is that turning off neurons in the first layers, which are responsible for learning basic features, prevents the network from extracting proper higher level features in subsequent layers.



\subsubsection{Disadvantages of projection-based methods}

Although projection-based methods have achieved meaningful performance, they still suffer from a severe information loss. As stated above, projections from 3D point clouds onto the 2D domain are very efficient in terms of complexity and inference time. However, these methods do not exploit the valuable 3D information and geometry. \autoref{fig:RvProj} illustrates that neighboring points that are located far away from each other with respect to L2-distance might be projected onto the same pixel in 2D space. Therefore, these methods fall short in terms of segmentation accuracy.

\subsection{3D-based methods}
Methods that keep the entire 3D information have been showing improved segmentation accuracy, in comparison to the previously presented works of the 2D domain, and have therefore become more popular recently. There exist several 3D approaches that process the data differently, i.e. voxel-based and point-based methods.

\subsubsection{Voxel-based methods}
\vspace{3mm} 
\textbf{Cylinder3D} \cite{zhou2020cylinder3d} proposes a novel voxel-grid partitioning method and integrates a new module into a 3D U-Net architecture \cite{3dUnet}. 

Instead of discretizing the space into regular voxel-grids, the authors proposed to perform cylinder partitioning, which is an encoding scheme that better represents the inherent distribution of point clouds. Cylinder3D is a voxel-based method, but it differs from classical methods of the same category, since the voxels are not shaped as regular cubes, but have a cylindrical shape instead. Similar to PolarNet, the points are transformed from Cartesian coordinates $(x,y,z)$ into the cylinder coordinate system $(\rho,\theta,z)$ based on the radius $\rho$ and azimuth $\theta$, with the difference that the height-dimension $z$ is preserved. To obtain the cylinder features for each voxel, the points are fed into a 4-layer MLP PointNet. After that, the maximum magnitude of the point features are selected as voxel representation.

Additionally, the authors introduce the Asymmetrical Residual Block that not only decreases the computational cost, but also better meets the property of cuboid objects. This module is used in all up-sampling and down-sampling blocks of the encoder-decoder network and is divided into two branches. The first branch consists of a $3\times1\times3$ convolution followed by a $1\times3\times3$ kernel, while the second branch is made up of the same filters in reverse order. Unlike a single $3\times 3\times 3$ convolution, stacking a $3\times1\times3$ and a $1\times3\times3$ filters covers the same receptive field with fewer parameters.
%
 
\subsubsection{Combined voxel-based and point-based methods}
Purely point-based methods suffer from exceedingly long runtimes
due to their complex local neighborhood search, as described in \autoref{sec:proj_based}. On top of that, they are not able to outperform projection-based methods regarding segmentation accuracy. Consequently, they are not considered in this work. However, some of the top performing approaches use a combination of both voxel-based and point-based methods to leverage the advantages of both feature representations. Commonly, the different features are processed on distinct branches, as outlined in \autoref{fig:SPVC}.

\vspace{3mm} 
\textbf{Sparse Point-Voxel Convolution (SPVC)}  \cite{tang2020searching} performs point-based and voxel-based operations on two separate branches, where the cost of communication between the two is negligible. The overall architecture is depicted in \autoref{fig:SPVC}.
Unlike sparse convolutional operations, the point-based branch can maintain a high resolution, which enhances the performance for smaller objects, such as pedestrians and bicyclists. It processes each point individually as in PointNet and therefore imposes only a minor additional computational burden on the overall architecture. In the upper branch, the point cloud is voxelized and further processed using sparse convolutions. The features are converted back to point-based representations through a trilinear interpolation that takes eight neighboring voxels into account. Finally, the information of both branches are fused with an addition.

\fig{90}{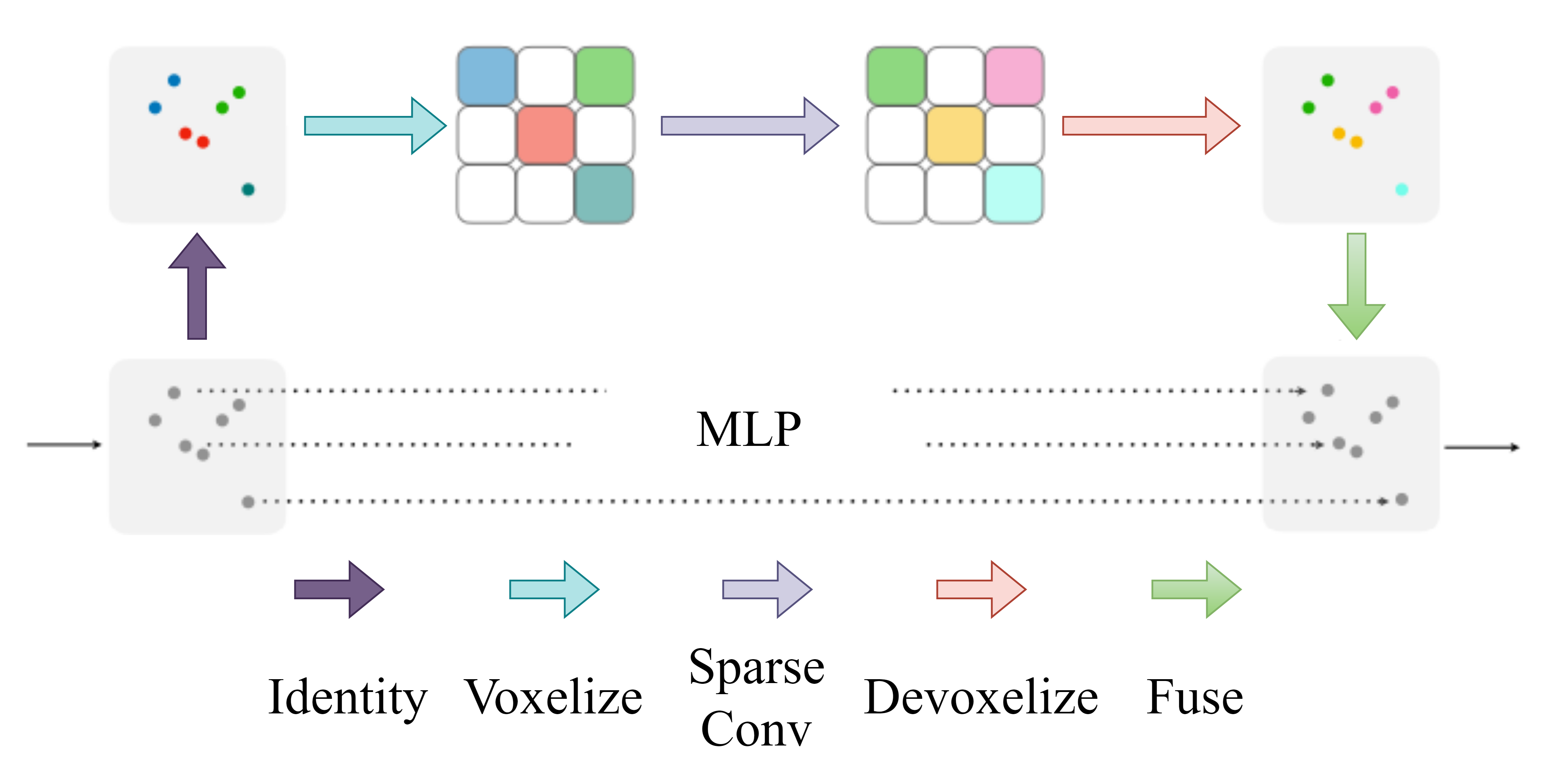}{\cite{tang2020searching} The SPVC architecture consists of two branches: A voxel-based branch to allow for large receptive fields and a point-wise branch to preserve geometrical information.}{SPVC}

Additionally, the authors introduce 3D Neural Architecture Search (3D-NAS) which is a framework that automatically explores a suitable architecture for the given application. While in some cases the only relevant factor is segmentation accuracy, most applications are constrained by latency or computational cost. To this end, a super network consisting of several SPVC networks (SPVNAS) with different channel number and network depth is trained, since it is infeasible to train each model with different hyperparameters independently. 

In each training iteration, different candidate networks with randomly selected channel number and network depth are sampled on different GPUs and the gradients are subsequently averaged. However, due to the fine-grained design space most network candidates are optimized only for a very few iterations and some may not be sampled at all. Thus, if all candidates had individual weights, they could not be trained sufficiently. To circumvent this problem, the authors employ the weight sharing technique during training. When a new candidate network is sampled with a depth $d_t > d_{t-1}$, then the first $d_{t-1}$ layers are kept, where $d_{t-1}$ is the depth of the previous sample. The same principle applies to the channel numbers $C_{in}$ and $C_{out}$. This ensures that each network is at least partially optimized at every iteration, even if it is not sampled. However, larger candidate networks are at a severe disadvantage as the deeper layer weights are barely trained. To address this problem, Progressive Depth Shrinking subdivides each training epoch into $m$ segments, where $m$ is the number of possible choices for the network depth $d$. During the $k^{th}$ segment of an epoch, $d$ must lie in the interval $[{m - k + 1},{m}]$.

After the training, Evolutionary Architecture Search is applied to determine the best possible model under the given resource constraints. To specify the resource constraints, the number of MACs (multiply-accumulate operations) is estimated based on the input and architecture, and is used as a threshold to narrow down the search space of possible candidate networks. 

\vspace{3mm} 
\textbf{DRINet++} \cite{ye2021drinet1}. 
Based on DRINet \cite{ye2021drinet}, this work performs feature learning by iteratively converting between voxel-based and point-based feature representations. DRINet++ extends this framework by introducing the concept of treating voxels as super points, which reduces memory consumption while increasing network performance. The idea is to process the raw point cloud into a voxel representation and to treat the voxels as points. Applying point-wise operations to voxels exploits the geometric properties of a point cloud. At the same time, each voxel can be interpreted as a combination of the features of neighboring points.  This significantly reduces the amount of memory-intensive point-wise operations that need to be performed.
%
%

The overall architecture mainly consists of two modules that pass on the learned features, Sparse Feature Encoder (SFE) and Sparse Geometric Feature Enhancement (SGFE), where each module takes the output of the other module as input, in an iterative process.

SFE takes a sparse voxel-based representation as input and performs sparse convolutions with a channel number of 64 in order to decrease computational overhead, while extracting local context information. The module uses a ResNet architecture with skip-connections \cite{resnet} and the Leaky-ReLU activation function.
In addition, Deep Sparse Supervision (DSS), an auxiliary loss function, is applied to the output of SFE in each iteration. This helps to significantly reduce memory consumption compared to dense supervision on each point. To reduce the inference time, the auxiliary loss is not calculated during the test phase.

The SGFE module implements the idea of "voxels as points", and is responsible for feature learning with more geometric guidance. It consists of two sub-blocks, where the sparse voxel-based features are first fed through a Multi-Scale Sparse Projection (MSP) block, which enhances predictive performance by making use of multi-scale features that can capture hierarchical geometry information of point clouds. Then, the output of the MSP block is further processed by Attentive Multi-scale Fusion (AMF).


Instead of fusing the multi-scale features naively based on predefined operations, the representations are fused with learnable weights in the AFM module. The reasoning behind this is that the significance of features from a certain scale highly depends on the input features.

The final prediction is obtained by applying nearest-neighbor interpolation to each point, taking the surrounding voxels into account. DSS is also applied to compute a loss for the prediction. However, at this stage DSS is not used as an auxiliary loss, and is therefore also computed during the testing phase.

\vspace{3mm} 
\textbf{\AFpaper} \cite{cheng2021af2s3net} is an encoder-decoder network with skip-connections that is based on sparse 3D convolutions from MinkNet42 \cite{choy20194d} and especially improves the segmentation performance on small objects. To this end, the architecture is extended with two novel modules, i.e. Multi-Branch Attentive Feature Fusion Module (AF2M) and Adaptive Feature Selection Module (AFSM). 

Before the input is fed into  the encoder, it is processed by AF2M. This module leverages three different kernel sizes ($3\times 3\times 3$, $4\times 4\times 4$ and $12\times 12\times 12$) on three branches that are based on both point-based and voxel-based features. This design choice helps to better extract contexts of different sizes and to extract both local details and global information. Every branch produces a $N\times 32$ feature map where $N$ is the number of points. Similar to the SGFE module in DRINet++, the features $\mathbf{x}_1$, $\mathbf{x}_2$ and $\mathbf{x}_3$ from the three branches are not merged naively but with an attention mechanism. To this end, the different features are fused by taking a weighted average of all different branches
\begin{equation}
    g\left(\mathbf{x}_{1}, \mathbf{x}_{2}, \mathbf{x}_{3}\right) = \alpha \mathbf{x}_{1}s+\beta \mathbf{x}_{2}+\gamma \mathbf{x}_{3}+\Delta.
\end{equation}
The weights $\alpha$, $\beta$, $\gamma$ are learnable masks. These are  obtained by a convolutional operation that reduces the depth of the corresponding feature maps $\mathbf{x}_1$, $\mathbf{x}_2$, $\mathbf{x}_3$ from 32 to 1. This is then further concatenated to a weight matrix of shape $N\times 3$ and a softmax operation is applied to the weight matrix to normalize each row (along the branch dimension) and to put more importance to certain features.  The attention residual $\Delta$ is supposed to stabilize the attention layer and obtained by applying another convolutional operation to the weight matrix. Therefore, the different branches of AF2M learn to put their attention to instances of different sizes. The fused features are finally convolved with another filter and passed to the encoder.

AFSM is a separate branch to the encoder-decoder network that takes the unweighted features $\mathbf{x}_1$, $\mathbf{x}_2$ and $\mathbf{x}_3$ as input. The purpose of this module is to filter out those feature maps that do not contribute to the final results. In this module the different features are further processed by convolutional units, concatenated, and are passed into a shared squeeze re-weighting network \cite{hu2019squeezeandexcitation} in which different feature maps are voted.

The output of AFSM is concatenated with the output of the last decoder layer. Then, another convolution is applied to reduce the depth of the feature map to the number of semantic classes.



\vspace{3mm} 

\textbf{JS3C-Net} \cite{yan2020sparse} aims to leverage the power of scene completion to enhance the predictive performance of a standard 3D U-Net architecture for Semantic Segmentation. To this end, the network architecture additionally consists of two auxiliary modules, i.e. Semantic Scene Completion (SSC) and Shape-aware Point-Voxel Interaction (PVI).


The first building block that performs Semantic Segmentation transforms $N$ points into a 3D voxel-grid by taking the average over all point features that fall into the same voxel. The input is then passed into a U-Net, that is based on submanifold sparse convolutions. The output of the network is transformed back to point-wise features using nearest-neighbor interpolation. Several MLPs are used to transform the point-wise features into a shape embedding $\mathbf{F}_{SE}$ and the per-class semantic probabilities $\mathbf{F}_{out}$ for all points.

The SSC module is introduced to tackle the problem of extreme sparsity in real-world 3D point clouds, where the data gets sparser with increasing distance to the LiDAR sensor. The input to the module is $\mathbf{F}_{out}$, the per-class semantic probabilities for every point.
First, $\mathbf{F}_{out}$ is voxelized and the resolution of the input is reduced by a pooling operation to decrease computational complexity. Then, it is passed through a series of convolutional layers with skip-connections. In the end, the features from different scales are concatenated and undergo a dense up-sampling \cite{Dense_Up_Sampling}. This is a continuous learning process that recovers the fine details which are typically lost in other naive up-sampling methods. The output of this module is the completed coarse voxelized scene, where each voxel holds one of the $C + 1$ labels (1 for non-objects class) and serve as contextual shape priors.

The PVI module performs knowledge fusion between SSC and the Semantic Segmentation block and helps both modules to mutually improve their performance. 
Since the SSC module operates on voxel-based feature maps, some information is inevitably lost. Therefore, the output of the module is passed on to the PVI module. There, it is fused with the shape embeddings $\mathbf{F}_{SE}$ from the Semantic Segmentation pipeline, which guides the coarse representation into a more fine-grained, completed scene. The center of each non-empty voxel coming from SCC is treated as a point and is connected to its k-nearest neighbors from the original point cloud in a graph convolutional network. The nodes of the graph consist of point features and point positions, and the initial edges are obtained with an MLP that takes the difference of the positions and features of two points into account. 

In order to reduce the computational burden during inference time, the SSC and PVI modules are used only during training. Learning this whole pipeline end-to-end and also optimizing the SSC module implicitly assists to refine the weights of the U-Net in the Semantic Segmentation module. 



\subsection{Hybrid methods}
In order to combine the advantages of fast inference time and higher segmentation accuracy, many works have been proposed that rely on both 3D-modules and projections to 2D. 

\vspace{3mm} 
\textbf{AMVNet} \cite{liong2020amvnet} leverages the advantage of range-view (RV) and bird's-eye-view (BEV) projections by using two projection-based networks, which can run in parallel during inference time. Furthermore, a late fusion approach with an assertion-guided sampling strategy is introduced, where only points that have been assigned different class labels from the two networks are considered. These points are further processed by a point head.

The first network is a RV-network that is fully convolutional and based  on a ResNet-like backbone with skip-connections. For down-sampling, strided convolutions are used. The input to the network is a $360^\circ$ LiDAR scan that is projected onto the image plane. After the last FCN layer, an additional RNN is introduced to help the RV-network learn spatial relationships of objects in the azimuth direction. For this purpose, the feature map $\mathbf{F}_{last}$ from the last layer is converted into a sequence of length $H\cdot W$ that consists of all spatial cells, where $W$ is the width and $H$ is the height of $\mathbf{F}_{last}$. Then, the sequence is fed into the RNN. Each input cell is of size $C$, corresponding to the depth of $\mathbf{F}_{last}$, and so is each output cell for each element of the sequence. Then, the output sequence is rearranged and stacked together to match the shape of $\mathbf{F}_{last}$. 

The second network is based on BEV-projections and follows the architecture of PolarNet.

Finally, the predictions of the two networks are merged in a late fusion step. To this end, the cosine similarity between the class predictions of both networks for all points is computed. If the cosine similarity is lower than a threshold, the two networks disagree and the corresponding point is flagged as uncertain. Otherwise, the final predictions are computed as $\sqrt{{\mathbf{f}_i}*{\mathbf{g}_i}}$, where ${\mathbf{f}_i}$ and ${\mathbf{g}_i}$ are the class predictions of both networks. 

The uncertain points are further processed by a 3D point head, which outputs the refined class predictions. The point head only contributes a negligible number of parameters to the overall architecture, which makes it extremely efficient. For every uncertain point, the authors extract point-level features of the point itself and of
the $N$ nearest neighbors. The features corresponding to the neighboring points consist of ${\mathbf{f}_i}$, ${\mathbf{g}_i}$ and the relative distance between the neighboring point and the uncertain point. They are independently processed by a 3-layer MLP and max-pooled. The output is concatenated with the point-level features consisting of ${\mathbf{f}_i}$, ${\mathbf{g}_i}$, the 3D point coordinates and the intensity of the uncertain point. This is further processed by FC-layers and gives a new output that contains the refined class predictions.


\vspace{3mm} 
\textbf{2D3DNet} \cite{genova2021learning} leverages predictions of a 2D Semantic Segmentation network to create a pseudo-ground-truth, that is in turn used to supervise the training of a 3D Segmentation model, without using any 3D labeled training data. 

Manual annotation of 3D data is rather expensive. Moreover, the different 3D datasets were acquired outdoors in different environments and with different sensor settings, so a 3D model that is trained on one dataset cannot generalize well to others. In contrast, annotated 2D Semantic Segmentation datasets, such as Cityscapes \cite{cordts2016cityscapes} or PASCAL VOC \cite{Everingham2014ThePV}, are larger and cover more diverse scenes. Furthermore, they do not suffer as much from different sensor configurations, which makes these models more generalizable to unseen data.
%
%
In order to exploit these advantages, the overall network is divided into three stages. Given a LiDAR point cloud and corresponding RGB images that are gathered simultaneously, the first stage is to label the 2D pixels using the state-of-the-art DeepLabv3 \cite{DeepLabv3} in 2D Semantic Segmentation as a backbone. 

Second, the LiDAR point cloud needs to be labeled in a multi-view fusion step.
Each point is projected onto the set of images to find a point-pixel correspondence with the best matching label. To find the best match, the goal is to discard uncertain labels of the data and keep only those with extremely high confidence. While the LiDAR data is collected at a rate of approximately 10 Hz, the 2D images are collected less frequently (2Hz). The asynchronous sensors lead to inconsistencies between the acquired data, especially when objects are moving at high speed. Therefore, a temporal filter is introduced that considers only the points acquired within a period of $\Delta t_{max} = 0.1$ seconds before or after the image was taken. In addition, 2D segmented scenes suffer from an occlusion problem, where pixels that represent specific objects, are segmented as part of others. Typically, these object differ significantly in terms of depth, e.g. a background building that is segmented as part of a foreground tree. Thus, an occlusion filter is used so that only correspondences of points and pixels that have a depth difference of less than $\tau_{max} = 1\%$ are approved. After the filtering process, the pixels are projected back onto the 3D point cloud, and undergo a weighted voting. The weights for each pixel are determined by their scores given to them by the mentioned filters, i.e. the smaller $\tau$ and $\Delta t$, the larger the weight.

Third, the actual 3D Segmentation model is trained using a network of 3D submanifold sparse convolution blocks, which is supervised by the created pseudo-ground-truth from the two previous steps. The input of the 3D model are the 3D points fused with one-hot encoded vectors that hold the class labels. However, these labels that are used as features cannot be generated in the same way as the pseudo-ground-truth. In this case, the input features and supervision would tend to agree at sparse supervision locations, and the model could not generalize to other data.

Therefore, a more relaxed voting scheme is used where $\tau_{max} $ is increased to $5\%$ and $\Delta t_{max}$ is increased to $10$ seconds. Instead of naively passing the correct labels through the network, the model learns to correct mistakes and to reason based on the geometry. Input features of points that do not pass the relaxed filters use the one-hot encodings from their nearest neighbors.

During test time, the 3D sparse convolutional model predicts the final class labels and a 2D Semantic Segmentation model can optionally be used to obtain the input features for the 3D model.





\begin{figure*}[htp]
    \centering
    \includegraphics[width=140mm,scale=1]{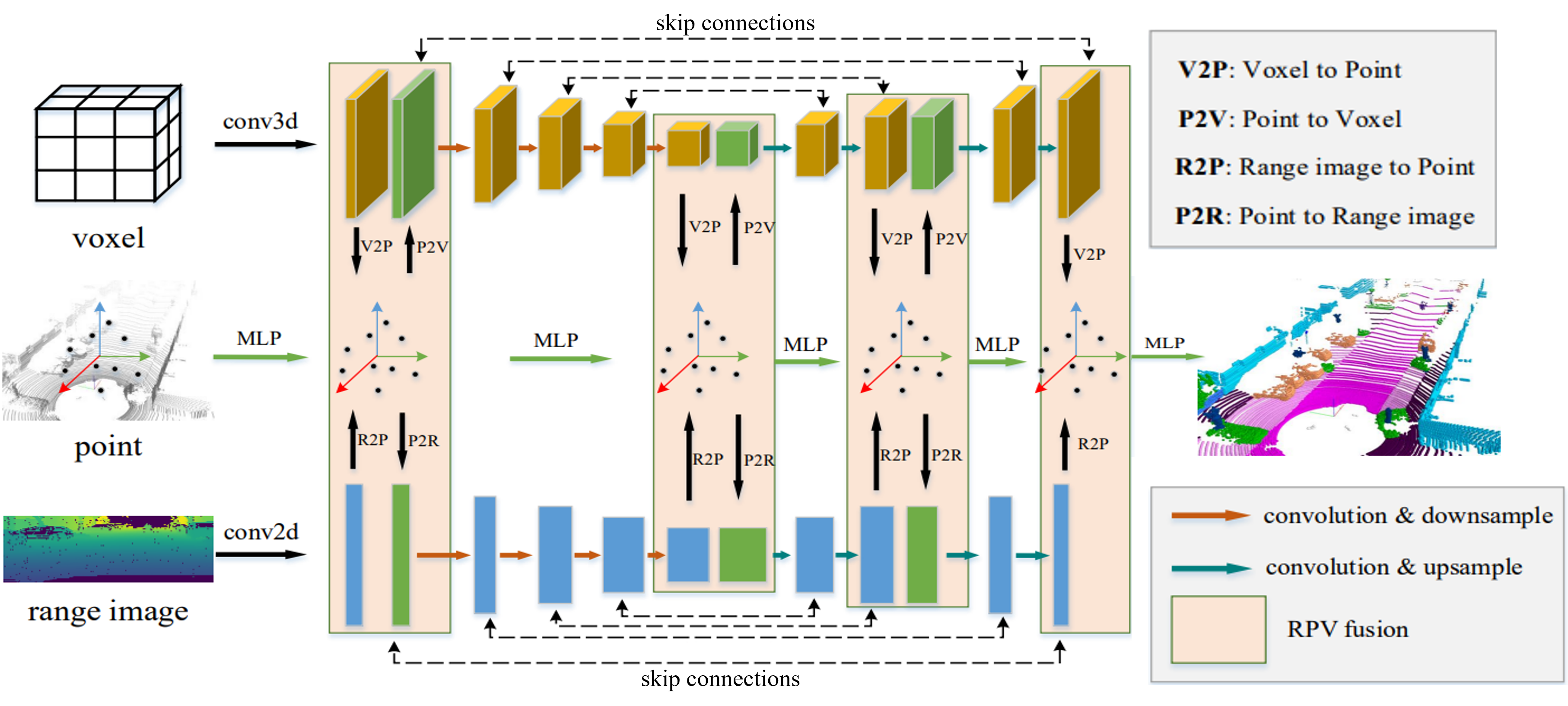}
    \caption{\cite{xu2021rpvnet}  Recent methods utilize multiple branches to exploit the advantages of different representations: This Figure illustrates the 3-branch architecture of RPV-Net. Both the voxelized and range-view representations are processed by U-Nets.}
    \label{fig:RPVnet}
\end{figure*}

\vspace{3mm} 
\textbf{RPVNet} \cite{xu2021rpvnet} exploits the advantages of point-based, voxel-based and projection-based methods on three different branches, as illustrated in \autoref{fig:RPVnet}.

Instead of fusing the different representations only once at the end, RPV-fusion is performed four times throughout the network so that the features of the three representations reinforce each other more frequently. 


The point representation is the central branch and serves as the mediator for feature fusion between the different branches. It uses a simple PointNet consisting of four MLPs which is extremely efficient because it does not perform the time-consuming search for local neighbors.

The other two branches follow a U-Net-like architecture which consists of four down-sampling and four up-sampling blocks. At the beginning of the U-Net, a stem is employed which performs convolutions without reducing the resolution, e.g. with a pooling operation, to extract more contextual information. To maintain efficiency, the voxel-based branch is based on sparse convolutions and the voxel resolution is set fairly low at 0.05 m.

The RVP-fusion between the three branches is divided into three stages. First, voxel-to-point (V2P) and range-to-point (R2P) operations are performed that transform both the voxel-based and the range-based features to a point representation. Every point has a hash code to search for the corresponding voxel or pixel more efficiently. V2P obtains a point representation with trilinear interpolation that takes eight neighboring voxels into account, and R2P uses bilinear interpolation based on four neighboring pixels. In the second stage, the three point representations are fused with a Gated Fusion Module (GFM). Instead of fusing the features naively with an addition or concatenation, a gating mechanism is used that determines the importance of each feature with learnable weights. This allows to avoid useless information and to prioritize some features over others. To this end, a weighted average of the three branches features is taken, where the weights sum up to 1. In the final step, the fused features of all $N$ points are projected back via the point-to-voxel (P2V) and point-to-range (P2R) operations. To this end, the voxel or pixel representation is obtained by averaging over all features that fall inside the same voxel or pixel. 

Furthermore, the Instance CutMix data augmentation technique is applied during training to tackle the issue of imbalanced classes. For this purpose, a mini sample pool is created that contains instances of classes that occur less frequently. During training, these objects are equally sampled by category, randomly scaled and rotated, and finally inserted in the current scene of the training. The objects are carefully placed only on top of ground-class points to ensure that the artificial changes in the scene are closely resembling real-world scenes.





\begin{table*}[h]
    \centering
    \makebox[\linewidth]{
    \begin{tabular}{l|cccccc}
        \midrule \midrule \text { Dataset } & \text { Enviroment} &\text { Scans } & \text { Points } & \text { Classes   } & \text { Category } \\
        \midrule \midrule 
        \text{ SemanticKITTI (2019) } \cite{behley2019semantickitti}  & \text {Karlsruhe} & 43,552 & 4.549 B & 28 &  \text { real } \\[0.1cm]
        \text { nuScenes-Lidarseg (2020) } \cite{caesar2020nuscenes}  & Boston, Singapore & 40,000 & 1.4 B & 32  & \text { real } \\[0.1cm]
        \text { SynLiDAR (2021) } \cite{xiao2021transfer}  & Unreal Engine 4 & 198,396 & 19.482 B & 32  & \text { synthetic } \\[0.1cm]
        \text { A2D2 (2020) } \cite{geyer2020a2d2}  & Southern Germany & 41,277 & - & 38 & \text { real } \\[0.1cm]
         \text { Paris-Lille-3D (2018) } \cite{roynard2018parislille3d}  & Paris, Lille & - & 143 M &  50 &  \text { real } \\[0.1cm]
        \text { Toronto-3D (2020) } \cite{Toronto-3D}  & Toronto & - & 78.3 M & 8 &  \text { real }\\
        \bottomrule 
    \end{tabular}}
    \caption{Overview of different autonomous driving datasets for 3D Semantic Segmentation.}
    \label{Table 1}
\end{table*}

\section{Evaluation}

\subsection{Datasets}
In recent years, a great deal of effort has gone into creating complete and meaningful datasets to support research on autonomous driving tasks in general and on Semantic Segmentation of scenes in particular. They allow researchers to focus on developing a novel model without having to collect and label vast amounts of data themselves. Additionally, datasets serve as benchmarks for comparing different approaches in 3D Semantic Segmentation.

\subsubsection{Real-world data for autonomous driving }


The SemanticKITTI and nuScenes datasets are the most important and extensive real-world datasets for Semantic Segmentation of point clouds and are therefore discussed subsequently.

\vspace{3mm} 
\textbf{SemanticKITTI} \cite{behley2019semantickitti} was collected around the city of Karlsruhe in Germany, and the scenes include highway scenarios, countryside roads, residential areas and inner city traffic.
This benchmark provides point-by-point semantic annotations of Velodyne HDL-64E LiDAR sensor point cloud data. It provides $23,201$ point clouds for training and $20,351$ for testing, along with 28 different semantic labels.


\vspace{3mm} 
\textbf{nuScenes} \cite{caesar2020nuscenes} is a large-scale autonomous driving dataset that was gathered in Boston and Singapore in a variety of different locations using six cameras, five Radar sensors and one LiDAR sensor. As of 2021, it contains $1.4$M 2D images, $390$K point clouds and $1.4$M bounding boxes for 32 different classes. 
The data was acquired at different times of the day and under different weather conditions, making it more diverse than the SemanticKITTI benchmark. The dataset was initially published for 3D Detection and Tracking, and was later extended by \textbf{nuScenes Lidarseg} which is used for 3D Semantic Segmentation on LiDAR data. As of now, nuScenes Lidarseg contains $1.4$ billion annotated points around 1000 different scenes where 850 scenes are dedicated to training and validation and the remaining 150 scenes are used for testing. 

\subsubsection{Synthetic data for autonomous driving }

A major issue with using real-world datasets is that manually labeling the data is extremely time-consuming. Also, these datasets are highly unbalanced, meaning that some object types are much more common than others. To address these issues, synthetic data can be used instead, where the point cloud is automatically labeled by creating authentic pseudo-ground-truth data. The main challenge is to overcome the domain gap between real-world data and synthetic data, as illustrated in \autoref{fig:PCT-net}. To this end, it is important to maintain a similar distribution and mimic the inherent properties and nature of LiDAR point clouds.

\begin{table*}
\setlength\tabcolsep{3pt}
\makebox[\linewidth]{
\begin{tabular}{c|c|cc|ccc|c}
\midrule \midrule
    & & & & \multicolumn{3}{c}{{SemanticKITTI}} \vline & nuScenes \\
\midrule \midrule
Catagory &Methods &  Parameters (M) & MACs (G) & Inference Time (ms) & mIoU ($\%$) & Hardware (GPU) & mIoU ($\%$)\\
\midrule 
\textbf{Projection} & \textbf{PolarNet} \cite{zhang2020polarnet} &  $13.6$ & $135.0$ & 62  &  57.2 & GTX1080Ti & 69.4 \\
&\textbf{SalsaNext} \cite{cortinhal2020salsanext} &  $6.7$ & $62.8$ & 71  &  59.5  & GTX1080Ti & -  \\
\midrule
\textbf{3D-based} & \textbf{Cylinder3D} \cite{zhou2020cylinder3d} & $53.3$ & $64.3$ & $131$  & $68.9$ & RTX 2080Ti & 77.2  \\
&\textbf{SPVNAS}(1) \cite{tang2020searching} &  $12.5$ & $73.8$ & 259  &  66.4 & GTX1080Ti & - \\
&\textbf{SPVNAS}(2) \cite{tang2020searching} &  1.1 & 8.9 & 89  &  60.3 & GTX1080Ti &  - \\
&\textbf{DRINet} \cite{ye2021drinet} & $3.5$ & $14.6$ & $62$ & $67.5$ & RTX 2080Ti & -  \\
&\textbf{DRINet}++ \cite{ye2021drinet1} &  $2.2$ & $12.1$ & \textbf{59}  & 70.7 & RTX 2080Ti  &  \textbf{80.4} \\
&\textbf{\small{\AFpaper}} \cite{cheng2021af2s3net} &  - & - & - &  69.7 &  Tesla V100  &  78.3 \\
&\textbf{JS3C-Net} \cite{yan2020sparse} & $2.7(+0.4)^{\star}$& - & $471(+107)^{\star}$ & 66.0 & Tesla V100  & 73.6   \\
\midrule
\textbf{Hybrid} & \textbf{AMVNet}  \cite{liong2020amvnet} &- & -& -& 65.3 & -& 77.3 \\
&\textbf{2D3DNet}  \cite{genova2021learning} & -& -& -& \textbf{72.7}& -& 80.0 \\
&\textbf{RPVNet}  \cite{ xu2021rpvnet} & $24.8$ & $119.5$ & $168$ & $70.3$ & RTX 2080Ti &  - \\
\midrule
\end{tabular}}
\caption{Comparison of the state-of-the-art methods for 3D Semantic Segmentation on the test sets of the SemanticKITTI and nuScenes benchmarks. The mIoU scores for nuScenes \cite{caesar2020nuscenes} are taken from the official benchmark. The remaining information is taken from the following sources: The numbers for  DRINet++ , DRINet, Cylinder3D and RPVNet are taken from \cite{ye2021drinet1}, the information for SalsaNext, PolarNet and SPVNAS is taken from \cite{tang2020searching} and the remaining information is taken from the papers themselves. $\star$ The additional numbers in brackets for JS3C-Net refer to the overhead of the SSC and PVI modules which can be discarded during inference. We furthermore provide two different configurations of SPVNAS with different resource constraints for a more accurate comparison and to demonstrate the effect of 3D-NAS.}
\label{table:SemanticKITTI comaprison}
\end{table*} 

\vspace{3mm} 
\textbf{SynLiDAR} \cite{xiao2021transfer}
is a highly comprehensive dataset of synthetic point cloud sweeps for 3D Semantic Segmentation that significantly exceeds existing real-world datasets in terms of both number of scans and points, as illustrated in \autoref{Table 1}.

The dataset consists of 13 LiDAR point cloud sequences with $\sim 200,000$ scans of point clouds with a total of $19$ billion points. It moreover offers 32 different semantic labels.

SynLiDAR is collected from a set of realistic virtual outdoor scenes created with the Unreal 4 graphics engine. The objects in the scenes are physically plausibly modeled by professional 3D generalists to ensure that the synthetic data accurately mimics the nature of real LiDAR point clouds in terms of both geometry and layout. In order to model the intensity values of the individual points in addition to the 3D coordinates, a rendering model is trained by learning from real LiDAR data.

In order to overcome the previously discussed domain gap, SynLiDAR is complemented with a point cloud translator called \textbf{PCT-Net}, that consists of two separate branches. The domain gaps are assumed to result primarily from differences in appearance due to variations between virtual and real scenes, or from differences in sparsity due to sensor sampling variations. In order to tackle both issues individually, the architecture consists of an Appearance Translation Module (ATM) and a Sparsity Translation Module (STM), as illustrated in \autoref{fig:PCT-net}.
Both branches take synthetic point clouds $X_s$ and real point clouds $X_r$ as input and are based on GANs. ATM generates a point cloud with real appearance information $X'_s$ and STM generates a point cloud with realistic sparsity $X''_s$, and finally $X'_s$ and $X''_s$ are fused together.

First, ATM performs up-sampling on $X_r$ and $X_s$ to disregard domain-specific sparsity information between them. Then, the generator takes the up-sampled $X_s$ and learns to output representations with real appearance that become indistinguishable from $X_r$ for the discriminator.

The STM module first projects $X_s$ and $X_r$ onto the image plane, since it is hard to capture sparsity information well in 3D space. A GAN based network takes the projected images $T(X_s)$ and $T(X_r)$ and performs image-to-image translation to ensure that $G_S(T(X_s))$ has similar sparsity as $T(X_r)$, where $G_S(.)$ is the operation performed by the generator of the STM module.

\fig{95}{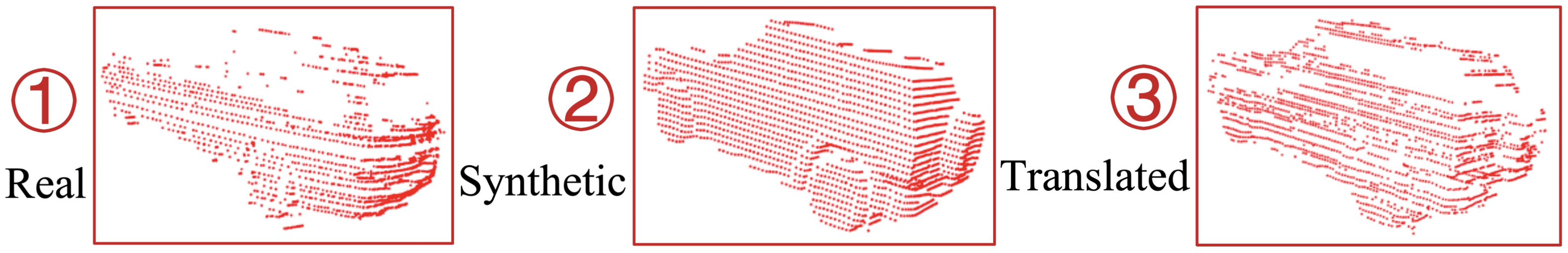}{\cite{xiao2021transfer} The domain gap between real-world data (1) and synthetic data (2). PCT-Net translates (2) to a more realistic representation (3) such that it mimics (1) well with respect to appearance and sparsity.}{PCT-net}

In addition, the SynLiDAR dataset addresses the problem of unbalanced distribution of objects by introducing a larger number of objects such as bicyclists and pedestrians, which are overshadowed by labels such as cars and vegetation in real-world datasets such as SemanticKITTI.

Experiments in \cite{xiao2021transfer} with this novel synthetic dataset demonstrate that it effectively augments and complements real-world datasets, and that it works well in a transfer learning setup.
Therefore, integrating SynLiDAR into the training process while using only real-world data during inference time results in better performance than standard data augmentation and increases the mIoU on SemanticKITTI by 2.2\% indicating that the data mimics real datasets well.

\subsection{Metrics}
 Mean Intersection over Union (mIoU) is used to evaluate the performance of 3D Semantic Segmentation models and is calculated as 
 \begin{equation}
     \text{mIoU} = \frac{1}{C} \sum_{c=1}^{C} \frac{\mathrm{TP}_{c}}{\mathrm{TP}_{c}+\mathrm{FP}_{c}+\mathrm{FN}_{c}},
 \end{equation}
where the average IoU over all semantic classes is taken. For every class $c$, the true positives ($\mathrm{TP}_{c}$) include all points where the class $c$ is predicted and agrees with the target class, i.e. the intersection. The denominator represents the union between prediction and target, where either both agree on the class $c$ ($\mathrm{TP}_{c}$), $c$ is incorrectly predicted ($\mathrm{FP}_{c}$), or the models fail to predict the correct class $c$ ($\mathrm{FN}_{c}$).

\section{Discussion}
In the following, interesting findings from the comparison of the selected state-of-the-art methods are discussed based on the details listed in \autoref{table:SemanticKITTI comaprison}. We selected 11 papers that propose the best performing approaches on the SemanticKITTI and nuScenes benchmarks. As for the performance, we do not only take into consideration the mIoU, but also the inference time. It is worth noting that all evaluations and comparisons between mIoU and inference time are discussed only for the SemanticKITTI benchmark. In addition, some novel, unpublished works that outperform the selected state-of-the-art methods, are already included in the leaderboards of the benchmarks. However, we only discuss works that have already published papers. It is also important to mention that not all results are perfectly comparable because the experiments were conducted on different hardware configurations.

In \autoref{table:SemanticKITTI comaprison} we compare the methods based on their efficiency. We observe that the number of parameters is not directly related to the inference time. Cylinder3D, for example, is very bulky with 53.3M parameters, but clearly outperforms JS3C-Net (2.7M parameters) in terms of inference time. We have also seen that some methods consist of multiple branches, which increases the number of parameters but not the runtime if inference can be done in parallel. 

As expected, the projection-based methods excel in terms of inference time. Still, the mIoU results are drastically lower than those of their 3D-based counterparts.
However, AMVNet combines the two different projection-based methods and thereby achieves an increase of more than 5$\%$ in mIoU, which points out that different types of projections complement each other.

Moreover, it is interesting to note that DRINet++, which is a 3D-based method, outperforms the projection-based methods in both mIoU scores and runtime. 
Nevertheless, all other approaches fail to outperform the projection-based methods with respect to inference time. In \autoref{table:SemanticKITTI comaprison}, we listed two different models of SPVNAS with different resource constraints. In the second version, the model size has been reduced to closely outperform SalsaNext in terms of mIoU. However, the model still has a slightly longer inference time, indicating that it cannot exceed projection-based methods in both mIoU and inference time simultaneously.

We can observe that the mIoU scores are higher on the nuScenes benchmark, which indicates that the test split of SemanticKITTI is more challenging.

In addition, 2D3DNet points out a major weakness of datasets for 3D semantic segmentation, as training a 3D model using only 2D supervision outperforms all other models trained on 3D data on the SemanticKITTI benchmark. This exhibits that current 3D datasets for Semantic Segmentation lag behind in terms of size and diversity. 

Furthermore, DRINet++ is the best method on the nuScenes benchmark in terms of mIoU. Similar to other methods that show superior mIoU scores, such as RPVNet and \AFpaper, this network merges point-based and voxel-based features. This indicates that merging multiple representations is very powerful.

\section{Conclusion and Future work}

In this review paper, we have provided a comprehensive overview of the current state-of-the-art methods in 3D Semantic Point Cloud Segmentation for the task of autonomous driving. While pioneering works in this area mainly investigated single feature representations (i.e. point-based, voxel-based, or projection-based), we have observed that the focus is shifting towards methods that combine different views. Leveraging multiple feature representations helps to increase the performance, which is a crucial step toward realizing the vision of an autonomous vehicle. 

However, the top performing approaches still do not exploit the power of synthetic datasets such as SynLiDAR  during training, which has been shown to improve their performance on real-world test data. 
 
Moreover, there are few approaches that investigate the use of temporal information from consecutive frames to address the problem of sparsity in point clouds. JS3C-Net has already explored the power of scene completion to densify point clouds. Another idea would be to fuse point clouds from multiple frames by finding point correspondences between different frames from a scene.


\bibliographystyle{IEEEtran}
\bibliography{References}

\end{document}